\documentclass[letterpaper, 10 pt, conference]{ieeeconf}
\IEEEoverridecommandlockouts
\usepackage{graphicx}
\usepackage{xcolor}
\usepackage{amsmath,amsfonts,amssymb}

\usepackage[T1]{fontenc}

\title{
Evaluating authenticity and quality of image captions via 
sentiment and semantic analyses \\
}

\author{Aleksei Krotov$^1$, Alison Tebo$^2$, Dylan K. Picart$^3$, Aaron Dean Algave
\thanks{*This work was facilitated by Neuromatch Academy DeepLearning course. Authors thank Joseph Akinyemi and Erum Afzal for their mentorship during coursework and project work.}
\thanks{$^{1}$Department of Bioengineering, Northeastern University, Boston, MA, $^{2}$HHMI - Janelia Research Campus, Ashburn, VA, $^{3}$Innovation Fellowship for Data Science, The Knowledge House, The Bronx, NY, 
{\tt\small bronze.eye@gmail.com, teboa@janelia.hhmi.org, dylankpicart@gmail.com, algaveaaron@gmail.com}}}%

\begin{document}
\maketitle

\begin{abstract}
The growth of deep learning (DL) relies heavily on huge amounts of labelled data for tasks such as natural language processing and computer vision. Specifically, in image-to-text or image-to-image pipelines, opinion (sentiment) may be inadvertently learned by a model from human-generated image captions. Additionally, learning may be affected by the variety and diversity of the provided captions. While labelling large datasets has largely relied on crowd-sourcing or data-worker pools, evaluating the quality of such training data is crucial.

This study proposes an evaluation method focused on sentiment and semantic richness. That method was applied to the COCO-MS dataset, comprising approximately 150K images with segmented objects and corresponding crowd-sourced captions. We employed pre-trained models (Twitter-RoBERTa-base and BERT-base) to extract sentiment scores and variability of semantic embeddings from captions. The relation of the sentiment score and semantic variability with object categories was examined using multiple linear regression. Results indicate that while most captions were neutral, about 6\% of the captions exhibited strong sentiment influenced by specific object categories. Semantic variability of within-image captions remained low and uncorrelated with object categories. Model-generated captions showed less than 1.5\% of strong sentiment which was not influenced by object categories and did not correlate with the sentiment of the respective human-generated captions. This research demonstrates an approach to assess the quality of crowd- or worker-sourced captions informed by image content.
\end{abstract}








\section{Introduction}

Deep learning (DL) research \cite{Hirota2022} has used a growing amount of worker-pooled or crowd-sourced labeling effort. Authenticity and quality of the such data are essential for training and evaluating the rapidly growing models in computer vision, natural language processing and other areas\cite{Chai2023, Ilyas2022}. However, systematic quality assessment of such data has remained limited \cite{Jorgensen1998,Ghandi2024}.  

In this study we proposed such evaluation of human-generated data labels, focusing on the positive-negative sentiment analysis and semantic variance across the raters. To demonstrate the proposed approach, we applied it to the COCO-MS dataset, using human-generated captions and segmented objects.
We next discuss the results which indicate a presence of strong sentiments in the captions, which were associated with the object categories, yet little semantic diversity across captions. 

\section{Methods}
We used the COCO-MS dataset consisting of about 150K images where 4-6 captions were sourced from data worker pool, and a thorough object segmentation was performed by a similar effort \cite{Lin2014}. 
Each caption was a single-sentence description of an image, and 80 object categories and 12 supercategories classified the segmented objects in the same image.

The captions were matched with the object categories via image ID. Sentiment analysis was performed on each caption using a pre-trained model Twitter-roBERTa-base. Sentiment score between -1 and 1 was computed as a difference between positive and negative confidence values \cite{Loureiro2022TimeLMS}. Strong sentiments were defined when the caption's score was larger than 0.5 or less than -0.5. The sentiment score was then examined against the presence or absence of object categories in the images. To that end, a subset of the data with strong sentiments was used. Influence of one-hot-encoded category presence on sentiment score was assessed via multiple linear regression (MLR) with first-order terms. 

Semantic analysis was performed on a set of the BERT-embeddings of the captions of each image. Variability was defined as the standard deviation of the cosine-similarity metric across the embeddings of the same image. To approximate that value without explicitly defining the mean in the cosine space, pairwise cosine similarities $S_c(a_i,a_j)$ for i-th and j-th captions were used in the following fashion: $s=\sqrt{\frac{1}{N(N-1)}\sum_{i=1}^{N} \sum_{j=1}^{N-1} S_c(a_i, a_j)^2}$

To examine a possible transfer of image-specific sentiment from the dataset to the model, the sentiment score was additionally compared to that of the model-generated captions for the same images. For producing captions, we used the Bootstrapping Language-Image Pre-training for Unified Vision-Language Understanding and Generation (BLIP) model for image captioning(\cite{https://doi.org/10.48550/arxiv.2201.12086}), which was trained on the COCO dataset. The train image dataset was used to generate captions with a minimum length of 8 words.

\section{Results}
\begin{figure}[t]
\centering
\vspace{-3mm}
\includegraphics[width=200px]{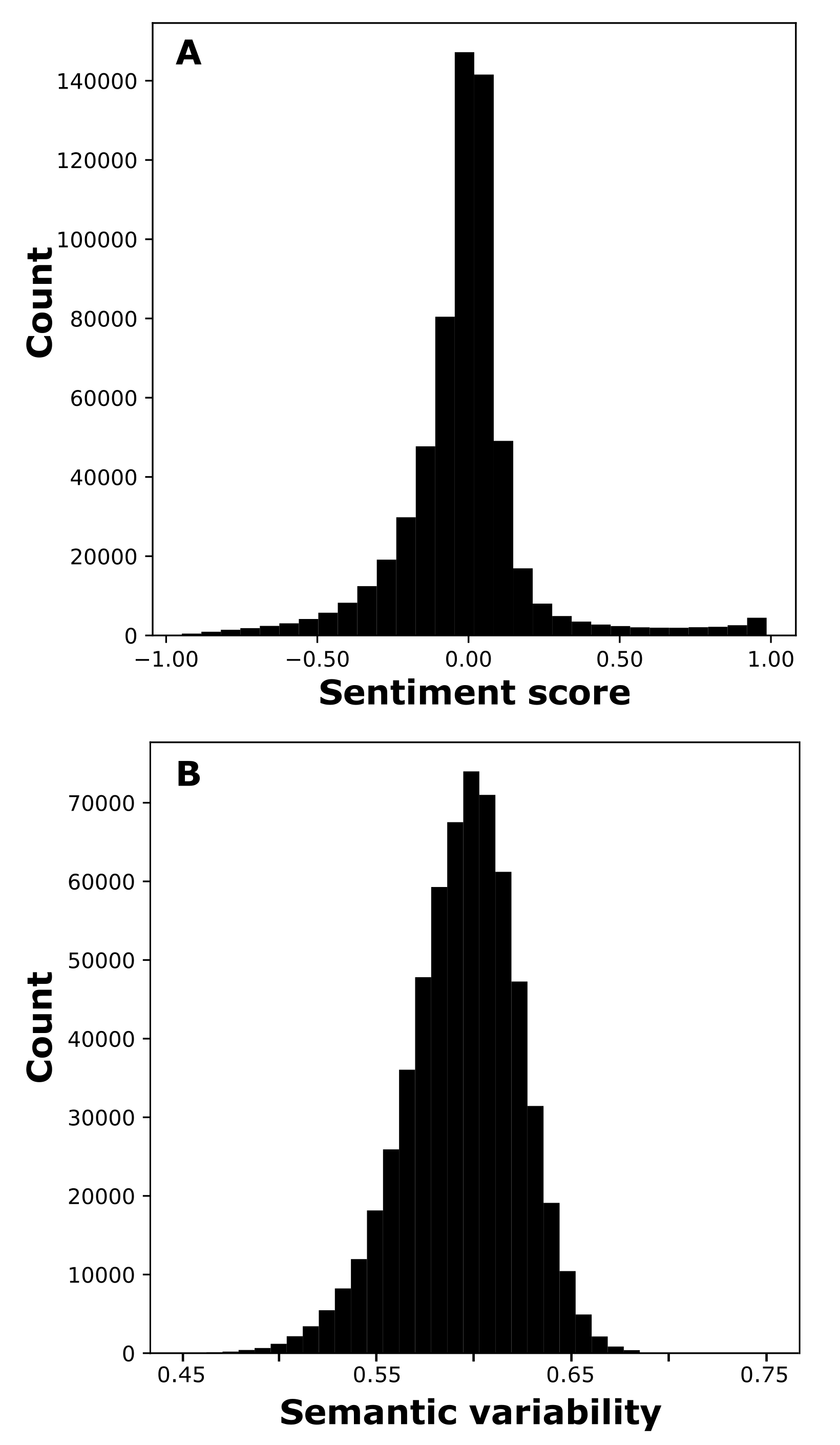}
\vspace{-3mm}
\caption{A: Distribution of Sentiment scores of human-generated captions of the COCO dataset. B: Distribution of Semantic variability of human-generated captions of the COCO dataset.}
\label{Figure-1}
\end{figure}

Most of the captions were found to carry a neutral sentiment (\ref{Figure-1}A), while about 6\% (32K) of the captions had either strong negative or strong positive sentiments, that correspond to 25K unique images. About 3.5K images had two strong captions, 800 had three strong captions, and 400 had three to five strong captions. The score mean and standard deviation per image did not appear to cluster in a two-dimensional space with respect to a category or a supercategory.
Unlike the long-tailed distribution of caption sentiments, the semantic variability within-image showed normal distribution with mean close to 0.6 (\ref{Figure-1}B). 


\begin{figure}[t]
\centering
\vspace{-3mm}
\includegraphics[width=200px]{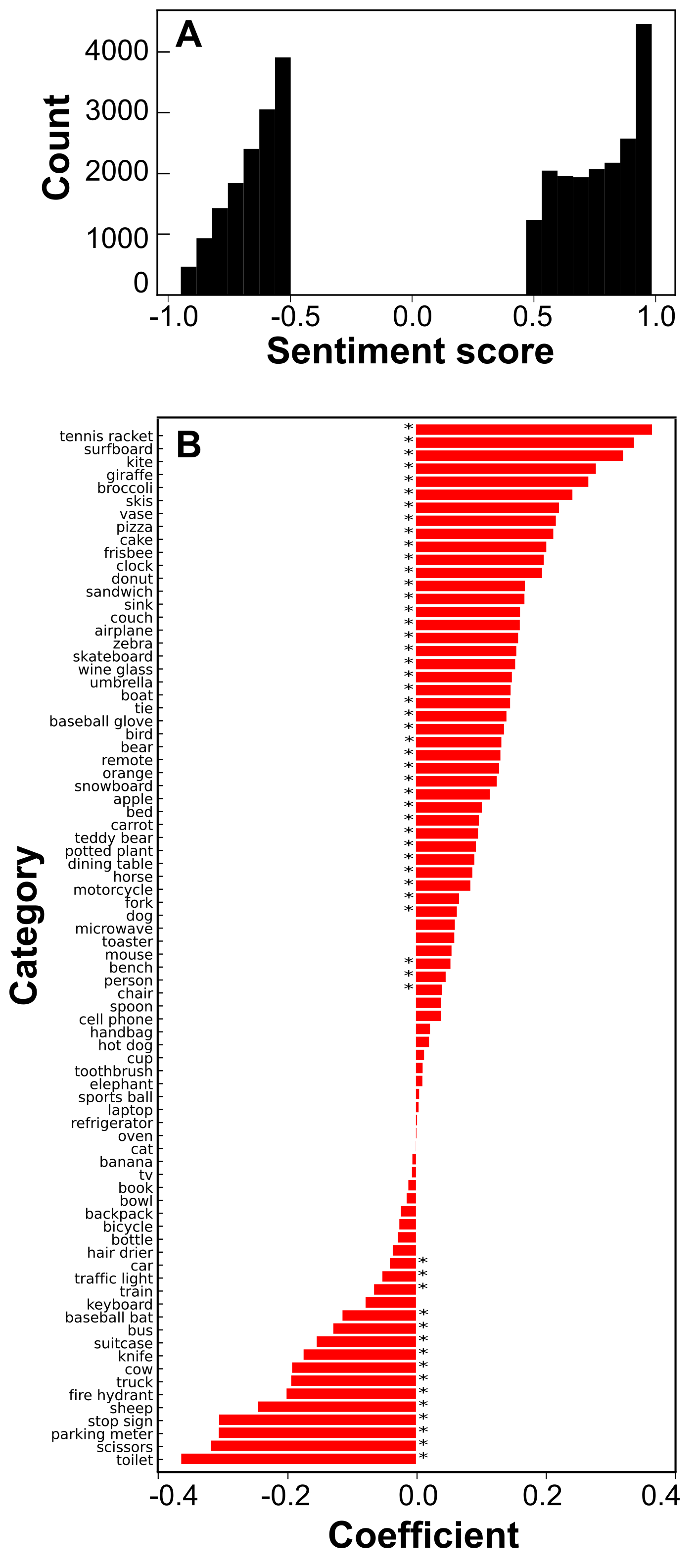}
\vspace{-3mm}
\caption{A: Distribution of Sentiment scores in the subset of strong-sentiment captions. B: Coefficients (slopes) of multiple linear regression of one-hot-encoded category vs. sentiment score. Asterisks mark coefficients significantly different from zero at p < 0.01}
\label{Figure-2}
\end{figure}

In the subset of the captions with strong sentiment (\ref{Figure-2}A), a linear model fit by MLR explained 24\% of the data variance and indicated that the presence or absence of most of the categories significantly affected the sentiment score. The results are summarized in \ref{Figure-2}B which shows the coefficient, or slope, of each one-hot-encoded category with respect to the sentiment score, and asterisks denote the coefficients which were significantly different from zero with \textit{p} < 0.01. 
Semantic variability and sentiment score showed no correlation, with Pearson's \textit{r} = 0.01.

\begin{figure}[tp]
\centering
\vspace{-6mm}
\includegraphics[width=185px]{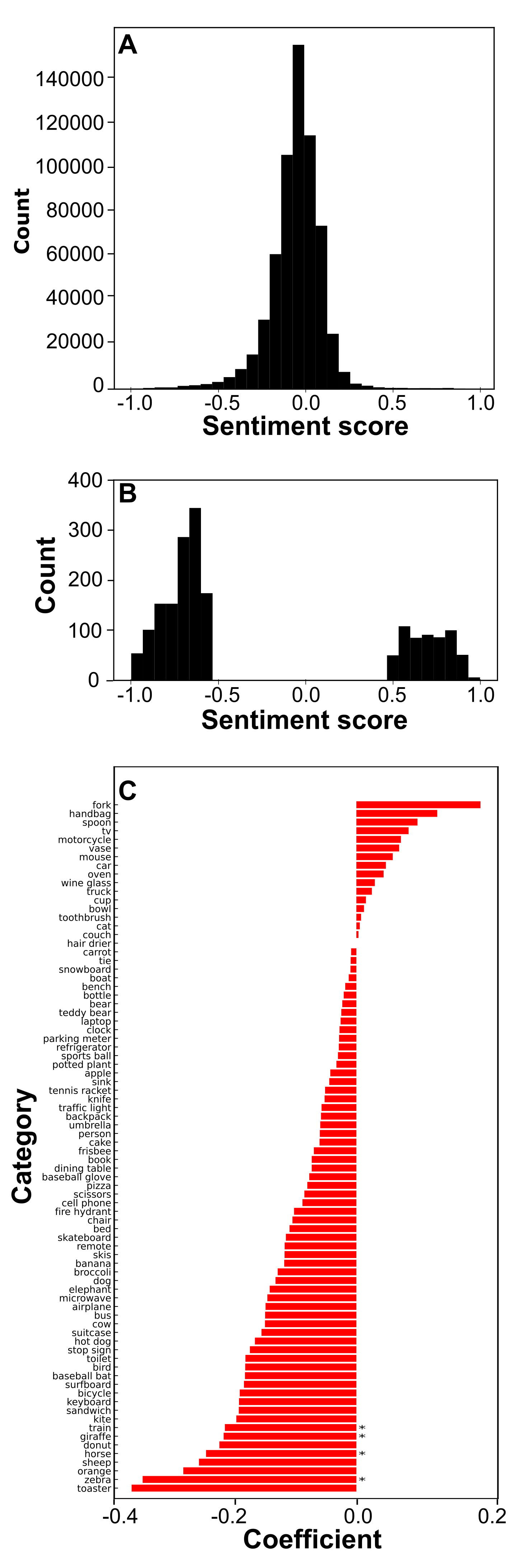}
\vspace{-3mm}
\caption{Distribution of Sentiment scores in model-generated captions, one per image.}
\label{Figure-3}
\end{figure}

Model-generated captions showed less-skewed distribution (\ref{Figure-3}A), than their human-generated counterparts. Only about 1.5\% of them (1800) indicated strong captions (\ref{Figure-3}B). The MLR fit to the subset of strong sentiments explained less than 13\% of their variance (\ref{Figure-3}C). Coefficient values corresponding to the contribution of each category's presence or absence resulted in a vastly different list from that in the human data, and only four of 80 were significantly different from zero.


\section{Discussion}
Nearly 10 years after the publication of the dataset MS-COCO, automatic caption generation shows high quality with respect to various aspects being investigated and available caption styles  \cite{Ghandi2024,klein2022diverseimagecaptioninggrounded}. However, existing approaches remain dependent on human-generated reference captions. This work investigated two aspects of such reference captions, sentiment and semantic richness, and their relation with objects in the image. 

Describing an image may be, in general, arbitrarily related to expressed sentiment. For objective captioning, however, weaker sentiment allows for conveying more factual information with less distraction. The workers providing captions for COCO dataset received instructions which promoted "describing important parts of the scene" but did not explicitly restrict or encourage sentiment expression \cite{Chen2015}. 
The caption relevance was then evaluated against some reference material via several semantic metrics (BLEU, Meteor, ROUGE and CIDEr) and via Precision at Human Recall, but was not seemingly inspected for expression of strong sentiment. Our study appears to be the first one addressing sentiment from the human-generating perspective, rather than evaluating model-generated instances.

The distribution of sentiment peaking at zero reflects an objective generation of captions. It shows however two slight asymmetries: a slight positive skew around center, and a high-positive tail. While examining the relationship between all the captions and object categories resulted in less than 10\% of variability explained, the object categories contributed to the captions with stronger sentiments, with a magnitude larger than 0.5. 
Research suggested that the sentiment of a whole image may be predicted by the sentiment of one of more salient objects in the image, such as faces, human-made objects, close-up views, or indoor scenes \cite{Zheng2017,Sun2016}. Our results suggest that human faces did not affect caption sentiments much (correlation with "person" is relatively low), but that other, non-human factors did.
The training-data sentiment distribution asymmetry and its relation with objects may not necessarily transfer to synthetic captions. The approach employed in this work resulted in a tighter and more symmetric distribution of caption sentiments and a lack of relation to object categories. It remains, however, an aspect for potential scrutiny in new generative language models. One way of addressing that issue may be focusing on strongest-sentiment object categories and collecting additional data with the same categories but opposite sentiments \cite{Mohammed2022}.

Our approach of using cosine similarity of sentence BERT-embeddings is an example of a rapid off-the-shelf metric to quantify semantic difference and scale it to semantic variability. Semantic variability of captions showed a symmetric and narrow distribution, mostly contained between 0.5 and 0.65, and a lack of relation with object categories. Compared to standard deviation, used as variability, cosine similarity between embeddings of a pair of captions is slightly larger on average. With that and with non-negative pairwise distances, the observed distribution of standard deviation reflects high similarity and low variability across captions of each image. This may arise from the use of common words between workers and the employment of similar or parallel sentence structures. While this indicates a good agreement between the workers, it also suggests low semantic diversity of the captions therefore, limiting the variation of the models trained on those data.

Research quantifying semantic diversity of human-sourced descriptions of situations suggests a relation with salient objects in a scene, which was not the case for our results \cite{Jorgensen1998,takmaz-etal-2024-describing,Parrigon2017}. One likely explanation is that a lower semantic diversity and lack of its object-dependence are specific to the population of workers who generated the data and the nature of their task environment and instructions.
Taken together, the results suggest a cautious approach to crowd-sourcing caption-like data to ensure its objectiveness and diversity. 

\bibliographystyle{IEEEtran}
\bibliography{References} 
\end{document}